\pdfoutput=1
\documentclass[11pt,letterpaper]{article}
\usepackage{emnlp2016}
\usepackage{times}
\usepackage{url}
\usepackage{booktabs}
\usepackage{multirow}
\usepackage{color}
\usepackage{graphicx}
\usepackage{latexsym}
\usepackage{nicefrac}
\usepackage{algorithm,algpseudocode}
\usepackage[colorinlistoftodos,prependcaption,textsize=small]{todonotes}

\usepackage{amsmath}
\usepackage{amsfonts,amssymb}
\usepackage[T1]{fontenc}

\makeatletter
\newcommand{\@BIBLABEL}{\@emptybiblabel}
\newcommand{\@emptybiblabel}[1]{}
\makeatother
\usepackage{hyperref}
\hypersetup{
    colorlinks,
    linkcolor={blue!25!black},
    citecolor={blue!25!black},
    urlcolor={blue!25!black}
}

\usepackage[activate={true,nocompatibility},final,tracking=true,kerning=true,spacing=true,factor=1100,stretch=10,shrink=10]{microtype}
\microtypecontext{spacing=nonfrench}

\newcommand{\pder}[2][]{\frac{\partial#1}{\partial#2}}
\newcommand{\given}{\, | \,}

\newcommand{\mbx}{\mathbf{x}}
\newcommand{\mbz}{\mathbf{z}}
\newcommand{\mbv}{\mathbf{v}}
\newcommand{\mbw}{\mathbf{w}}

\newcommand{\E}{\mathbb{E}}

\DeclareRobustCommand{\KL}[2]{\ensuremath{\textrm{KL}\left(#1\;\|\;#2\right)}}

\usepackage[acronym,smallcaps,nowarn]{glossaries}
\glsdisablehyper
\makeglossaries
\newacronym{EM}{em}{Expectation-Maximization}
\newacronym{NEM}{nem}{Neuralized Expectation Maximization}
\newacronym{ELBO}{elbo}{evidence lower bound}
\newacronym{KL}{kl}{Kullback-Leibler}
\newacronym{VI}{vi}{Variational Inference}
\emnlpfinalcopy

\def\emnlppaperid{10}

\title{Unsupervised Neural Hidden Markov Models}

\author{Ke Tran$^2$\Thanks{\ This research was carried out while all authors were at the Information Sciences Institute.}\ \ \ \ Yonatan Bisk$^1$\ \ \ Ashish Vaswani$^3$\footnotemark[1]\ \ \ \ Daniel Marcu$^1$\ \ \ Kevin Knight$^1$\\  
$^{1}$Information Sciences Institute, University of Southern California\\
$^{2}$Informatics Institute, University of Amsterdam\\
$^{3}$Google Brain, Mountain View\\
\href{mailto:m.k.tran@uva.nl}{\nolinkurl{m.k.tran@uva.nl}},
\href{mailto:ybisk@isi.edu}{\nolinkurl{ybisk@isi.edu}},\\ \href{mailto:avaswani@google.com}{\nolinkurl{avaswani@google.com}},
\href{mailto:marcu@isi.edu}{\nolinkurl{marcu@isi.edu}}, \href{mailto:knight@isi.edu}{\nolinkurl{knight@isi.edu}}
}

\date{}

\begin{document}

\maketitle

\begin{abstract}
In this work, we present the first results for neuralizing an Unsupervised Hidden Markov Model.  We evaluate our approach on tag induction.  Our approach outperforms existing generative models and is competitive with the state-of-the-art though with a simpler model easily extended to include additional context.
\end{abstract}

\section{Introduction}
Probabilistic graphical models are among the most important tools available to the NLP community.  In particular, the ability to train generative models using Expectation-Maximization (EM), Variational Inference (VI), and sampling methods like MCMC has enabled the development of unsupervised systems for tag and grammar induction, alignment, topic models and more.  These latent variable models discover hidden structure in text which aligns to known linguistic phenomena and whose clusters are easily identifiable.  

Recently, much of supervised NLP has found great success by augmenting or replacing context, features, and word representations with embeddings derived from Deep Neural Networks.  These models allow for learning highly expressive non-convex functions by simply backpropagating prediction errors. Inspired by \newcite{berg2010painless}, who bridged the gap between supervised and unsupervised training with features, we bring neural networks to unsupervised learning by providing evidence that even in unsupervised settings, simple neural network models trained to maximize the marginal likelihood can outperform more complicated models that use expensive inference.  

In this work, we show how a single latent variable sequence model, Hidden Markov Models (HMMs), can be implemented with neural networks by simply optimizing the incomplete data likelihood.  The key insight is to perform standard forward-backward inference to compute posteriors of latent variables and then backpropagate the posteriors through the networks to maximize the likelihood of the data.

Using features in unsupervised learning has been a fruitful enterprise \cite{das-petrov:2011:ACL-HLT2011,bergkirkpatrick-klein:2010:ACL,cohen-das-smith:2011:EMNLP} and attempts to combine HMMs and Neural Networks date back to 1991 \cite{bengio:1991}. Additionally, similarity metrics derived from word embeddings have also been shown to improve unsupervised word alignment~\cite{songyot2014improving}.  

Interest in the interface of graphical models and neural networks has grown recently as new inference procedures have been proposed \cite{Kingma:2013tz,Johnson:2016ud}.  Common to this work and ours is the use of neural networks to produce potentials. The approach presented here is easily applied to other latent variable models where inference is tractable and are typically trained with EM. We believe there are three important strengths:
\begin{enumerate}
\item Using a neural network to produce model probabilities allows for seamless integration of additional context not easily represented by conditioning variables in a traditional model.
\item Gradient based training trivially allows for multiple objectives in the same loss function. 
\item Rich model representations do not saturate as quickly and can therefore utilize large quantities of unlabeled text.
\end{enumerate}

Our focus in this preliminary work is to present a generative neural approach to HMMs and demonstrate how this framework lends itself to modularity (e.g. the easy inclusion of morphological information via Convolutional Neural Networks \S \ref{sec:conv}), and the addition of extra conditioning context (e.g. using an RNN to model the sentence \S \ref{sec:lstm}).  Our approach will be demonstrated and evaluated on the simple task of part-of-speech tag induction. Future work, should investigate the second and third proposed strengths.

\section{Framework}
\label{sec:framework}
Graphical models have been widely used in NLP. Typically potential functions $\psi(\mbz, \mbx)$ over a set of latent variables, $\mbz$, and observed variables, $\mbx$, are defined based on hand-crafted features. Moreover, independence assumptions between variables are often made for the sake of tractability. Here, we propose using neural networks (NNs) to produce the potentials since neural networks are universal function approximators. Neural networks can extract useful task-specific abstract representations of data.  Additionally, Long Short-Term Memory (LSTM) \cite{Hochreiter:1997} based Recurrent Neural Networks (RNNs), allow for modeling unbounded context with far fewer parameters than naive one-hot feature encodings. The reparameterization of potentials with neural networks (NNs) is seamless:
\begin{align}
\psi(\mbz, \mbx) &= f_{\text{NN}}(\mbz, \mbx \given \theta)
\end{align}

The sequence of observed variables are denoted as $\mbx = \{x_1, \dots, x_n\}$. In unsupervised learning, we aim to find model parameters $\theta$ that maximize the evidence $p(\mbx \given \theta)$.  We focus on cases when the posterior is tractable and we can use Generalized EM \cite{Dempster:1977ul} to estimate $\theta$. 
\begin{align}
p(\mbx) &= \sum_z p(\mbx, \mbz)\\
	&= \E_{q(\mbz)}[\ln p(\mbx, \mbz \given \theta)] + \mathrm{H}[q(\mbz)]\\
    &+ \KL{q(\mbz)}{p(\mbz\given\mbx,\theta)}
\label{eq:DL}
\end{align}
where  $q(\mbz)$ is an arbitrary distribution, and $\mathrm{H}$ is the entropy function. The E-step of EM  estimates the posterior $p(\mbz\given\mbx)$ based on the current parameters $\theta$.  In the M-step, we choose $q(\mbz)$ to be the posterior $p(\mbz\given\mbx)$, setting the KL-divergence to zero.  Additionally, the entropy term $\mathrm{H}[q(\mbz)]$ is a constant and can therefore be dropped.  This means updating $\theta$ only requires maximizing $\E_{p(\mbz\given\mbx)}[\ln p(\mbx, \mbz \given \theta)]$.  The gradient is therefore defined in terms of the gradient of the joint probability scaled by the posteriors:
\begin{align}
J(\theta) &= \sum_\mbz p(\mbz\given\mbx) \pder[\ln p(\mbx, \mbz\given\theta)]{\theta}
\label{eq:gradient}
\end{align}

In order to perform the gradient update in Eq~\ref{eq:gradient}, we need to compute the posterior $p(\mbz\given\mbx)$. This can be done efficiently with the Message Passing algorithm. Note that, in cases where the derivative $\pder{\theta}\ln p(\mbx, \mbz\given\theta)$ is easy to evaluate, we can perform direct marginal likelihood optimization \cite{salakhutdinov:2003}. We do not address here the question of semi-supervised training, but believe the framework we present lends itself naturally to the incorporation of constraints or labeled data.  Next, we demonstrate the application of this framework to HMMs in the service of part-of-speech tag induction.

\section{Part-of-Speech Induction}
Part-of-speech tags encode morphosyntactic information about a language and are a fundamental tool in downstream NLP applications.  In English, the Penn Treebank \cite{Marcus:1994tg} distinguishes 36 categories and punctuation.  Tag induction is the task of taking raw text and both discovering these latent clusters and assigning them to words in situ.  Classes can be very specific (e.g. six types of verbs in English) to their syntactic role.  Example tags are shown in Table \ref{tab:tags}.  In this example, \textit{board} is labeled as a singular noun while \textit{Pierre Vinken} is a singular proper noun.

\begin{table}
\centering
\begin{small}
\begin{tabular}{@{}lc@{\hspace{8pt}}c@{\hspace{8pt}}c@{\hspace{8pt}}c@{\hspace{8pt}}c@{\hspace{8pt}}c@{}}
\textbf{Text} & Pierre & Vinken & will & join & the & board\\
\textbf{PTB} & \texttt{NNP} & \texttt{NNP} & \texttt{MD} & \texttt{VB} & \texttt{DT} & \texttt{NN}\\
\end{tabular}
\end{small}
\caption{Example Part-of-Speech tagged text.}
\label{tab:tags}
\end{table}

Two natural applications of induced tags are as the basis for grammar induction \cite{spitkovsky-EtAl:2011:EMNLP,bisk-christodoulopoulos-hockenmaier:2015:ACL-IJCNLP} or to provide a syntactically informed, though unsupervised, source of word embeddings.  

\subsection{The Hidden Markov Model}
A common model for this task, and our primary workhorse, is the Hidden Markov Model trained with the unsupervised message passing algorithm, Baum-Welch \cite{Welch:2003wp}. 
\paragraph{Model}

\begin{figure}
\centering
\includegraphics[width=0.8\linewidth]{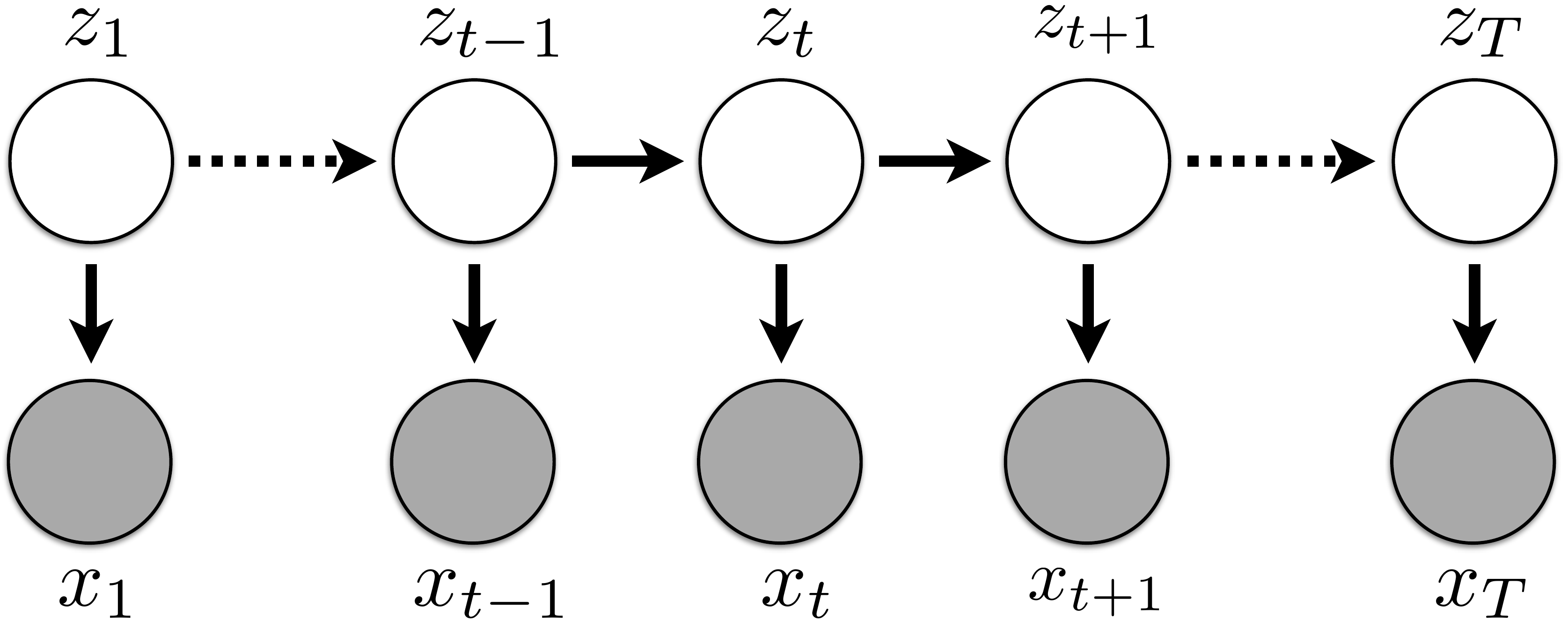}
\caption{Pictorial representation of a Hidden Markov Model.  Latent variable ($z_t$) transitions  depend on the previous value ($z_{t-1}$), and emit an observed word ($x_t$) at each time step.}
\label{fig:HMM}
\end{figure}

HMMs model a sentence by assuming that (a) every word token is generated by a latent class, and (b) the current class at time $t$ is conditioned on the local history $t-1$. Formally, this gives us an emission $p(x_t \given z_t)$ and transition $p(z_t \given z_{t-1})$ probability.  The graphical model is drawn pictorially in Figure \ref{fig:HMM}, where shaded circles denote observations and empty ones are latent.  The probability of a given sequence of observations $\mathbf{x}$ and latent variables $\mathbf{z}$ is given by multiplying transitions and emissions across all time steps (Eq. \ref{eq:HMM}).  Finding the optimal sequence of latent classes corresponds to computing an argmax over the values of $\mathbf{z}$.

\begin{equation}
\displaystyle p(\mbx, \mbz) =  \prod_{t=1}^{n+1} p(z_t\given z_{t-1}) \prod_{t=1}^{n} p(x_t \given z_t)
\label{eq:HMM}
\end{equation}

Because our task is unsupervised we do not have a priori access to these distributions, but they can be estimated via Baum-Welch.  The algorithm's outline is provided in Algorithm \ref{alg:BW}.

\begin{algorithm}
\caption{Baum-Welch Algorithm}
\label{alg:BW}
\begin{algorithmic}
\State Randomly Initialize distributions ($\theta$)
\Repeat
\State Compute forward messages: \hfill $\forall_{i,t} \; \alpha_i(t)$
\State Compute backward messages: \hfill $\forall_{i,t} \; \beta_i(t)$
\State Compute posteriors: \\
\hspace{3em} $p(z_t = i \given \mbx, \theta)  \propto \alpha_i(t)\beta_i(t)$ \\ 
\hspace{3em} $p(z_t = i, z_{t+1} = j \given \mbx, \theta)$\\ 
\hspace{3em} 	\quad\quad \quad  $\propto \alpha_i(t) p(z_{t+1}\!=\!j|z_{t}\!=\!i)$\\
\hspace{3em} 	\quad\quad  $\quad \quad \times \beta_j(t+1) p(x_{t+1}|z_{t+1}\!=\! j)$
\State Update $\theta$
\Until{Converged}
\end{algorithmic}
\end{algorithm}

Training an HMM with EM is highly non-convex and likely to get stuck in local optima
\cite{Johnson:2007tg}.  Despite this, sophisticated Bayesian smoothing leads to state-of-the-art performance \cite{blunsom-cohn:2011:ACL-HLT2011}.  \newcite{blunsom-cohn:2011:ACL-HLT2011} further extend the HMM by augmenting its emission distributions with character models to capture morphological information and a tri-gram transition matrix which conditions on the previous two states.  Recently, \newcite{lin-EtAl:2015:NAACL-HLT} extended several models including the HMM to include pre-trained word embeddings learned by different skip-gram models.  Our work will fully neuralize the HMM and learn embeddings during the training of our generative model.  There has also been recent work on by \newcite{rastogi-cotterell-eisner:2016:N16-1} on neuralizing Finite-State Transducers. 

\subsection{Additional Comparisons}
While the main focus of our paper is the seamless extension of an unsupervised generative latent variable model with neural networks, for completeness we will also include comparisons to other techniques which do not adhere to the generative assumption.  We include Brown clusters \cite{Brown:1992wr} as a baseline and two clustering techniques as state-of-the-art comparisons: \newcite{Christodoulopoulos:2011ud} and \newcite{yatbaz-sert-yuret:2012:EMNLP-CoNLL}.

Of particular interest to us is the work of \newcite{Brown:1992wr}.  Brown clusters group word types through a greedy agglomerative clustering according to their mutual information across the corpus based on bigram probabilities.  Brown clusters do not account for a word's membership in multiple syntactic classes, but are a very strong baseline for tag induction.  It is possible our approach could be improved by augmenting our objective function to include mutual information computations or a bias towards a harder clustering.


\section{Neural HMM}
The aforementioned training of an HMM assumes access to two distributions: (1) Emissions with $K\times V$ parameters, and (2) Transitions with $K\times K$ parameters.  Here we assume there are $K$ clusters and $V$ word types in our vocabulary.  Our neural HMM (NHMM) will replace these matrices with the output of simple feed-forward neural networks. All conditioning variables will be presented as input to the network and its final softmax layer will provide probabilities.  This should replicate the behavior of the standard HMM, but without an explicit representation of the necessary distributions.

\subsection{Producing Probabilities}
Producing emission and transition probabilities allows for standard inference to take place in the model.

\paragraph{Emission Architecture}
Let $\mbv_k \in \mathbb{R}^D$ be vector embedding of tag $z_k$, $\mbw_i\in\mathbb{R}^D$ and $b_i$ vector embedding and bias of word $i$ respectively. The emission probability $p(w_i\given z_k)$ is given by
\begin{align}
p(w_i\given z_k) &= \frac{\exp(\mbv_k^\top\mbw_i + b_i)}{\sum_{j=1}^V \exp(\mbv_k^\top\mbw_j + b_j)}
\end{align}
The emission probability can be implemented by a neural network where $\mbw_i$ is the weight of unit $i$ at  the output layer of the network. The tag embeddings $\mbv_k$ are obtained by a simple feed-forward neural network consisting of a lookup table following by a non-linear activation (ReLU). When using morphology information (\S\ref{sec:conv}), we will first use another network to produce the word embedddings $\mbw_i$.

\paragraph{Transition Architecture}
We produce the transition probability directly by using a linear layer of $D\times K^2$. More specifically, let $\mathbf{q} \in \mathbb{R}^D$ be a \emph{query embedding}. The unnormalized transition matrix $\mathbf{T}$ is computed as
\begin{align}
\mathbf{T} &= \mathbf{U}^\top\mathbf{q} + \mathbf{b}
\end{align}
where $\mathbf{U}\in\mathbb{R}^{D\times K^2}$ and $\mathbf{b}\in \mathbb{R}^{K^2}$. We then reshape $\mathbf{T}$ to a $K\times K$ matrix and apply a softmax layer per row to produce valid transition probabilities.

\subsection{Training the Neural Network}
The probabilities can now be used to perform the aforementioned forward and backward passes over the data to compute posteriors.  In this way, we perform the E-step as though we were training a vanilla HMM.
Traditionally, these values would simply be re-normalized during the M-step to re-estimate model parameters. Instead, we use them to re-scale our gradients (following the discussion from \S \ref{sec:framework}). Combining the HMM factorization of the joint probability $p(\mathbf{x},\mathbf{z})$ from Eq.~\ref{eq:HMM} with the gradient from Eq.~\ref{eq:gradient}, yields the following update rule:
\begin{align}
J(\theta) &= \sum_{\mbz} p(\mbz\given\mbx) \pder[\ln p(\mbx, \mbz\given\theta)]{\theta} \nonumber\\
& = \sum_t\sum_{z_t} p(z_t\given \mbx) \pder[\ln p(x_t\given z_t,\theta)]{\theta}  \nonumber\\
&\quad + p(z_t,z_{t-1}\given \mbx) \pder[\ln p(z_t\given z_{t-1},\theta)]{\theta}
\label{eq:hmmgrad}
\end{align}
The posteriors $p(z_t\given \mbx)$ and $p(z_t,z_{t-1}\given \mbx)$ are obtained by running Baum-Welch as shown in Algorithm~\ref{alg:BW}.
Where traditional supervised training can follow a clear gradient signal towards a specific assignment, here we are propagating the model's (un)certainty instead.  An additional complication introduced by this paradigm is the question of how many gradient steps to take on a given minibatch.  In incremental EM the posteriors are simply accumulated and normalized.  Here, we repeatedly recompute gradients on a minibatch until reaching the maximum number of epochs or a convergence threshold is met.  

Finally, notice that the factorization of the HMM allows us to evaluate the joint distribution $p(\mbx,\mbz\given\theta)$ easily. We therefore employ Direct Marginal Likelihood (DML) \cite{salakhutdinov:2003} to optimize the model's parameters.  After trying both EM and DML we found EM to be slower to converge and perform slightly weaker.  For this reason, the presented results will all be trained with DML.

\subsection{HMM and Neural HMM Equivalence}
An important result we see in Table \ref{tab:English} is that the Neural HMM (NHMM) performs almost identically to the HMM.  At this point, we have replaced the underlying machinery, but the model still has the same information bottlenecks as a standard HMM, which limit the amount and type of information carried between words in the sentence.  Additionally, both approaches are optimizing the same objective function, data likelihood, via the computation of posteriors.  The equivalency is an important sanity check.  The following two sections will demonstrate the extensibility of this approach.

\section{Convolutions for Morphology}
\label{sec:conv}
\begin{figure}
\centering
\includegraphics[width=0.8\linewidth]{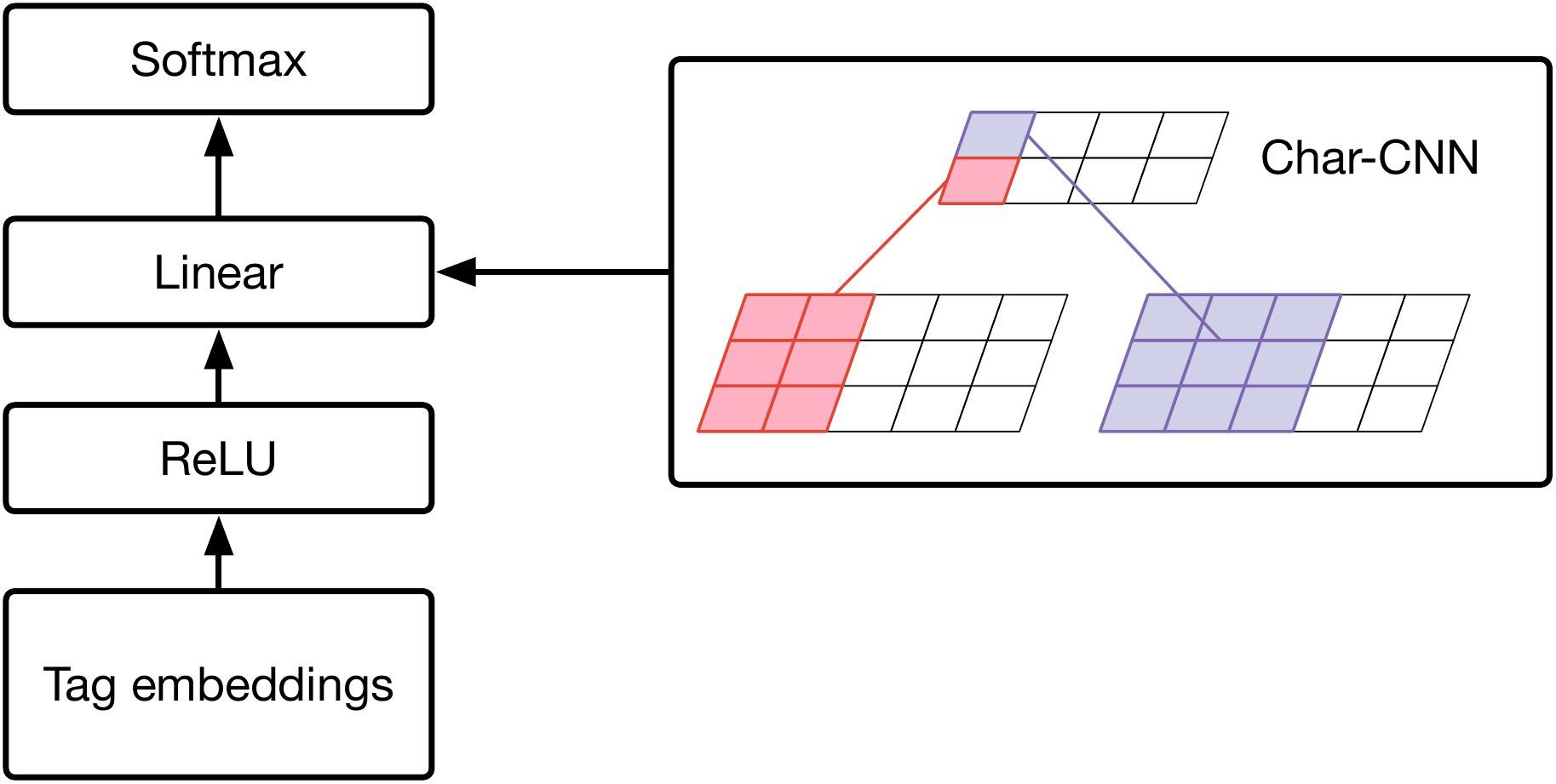}
\caption{Computational graph of Char-CNN emission network. A character convolutional neural network is used to compute the weight of the linear layer for every minibatch.}
\label{fig:charcnn}
\end{figure}

The first benefit of moving to neural networks is the ease with which new information can be provided to the model.  The first experiment we will perform is replacing words with embedding vectors derived from a Convolutional Neural Network (CNN) \cite{Kim:2015cnnlstm,Jozefowicz:2016}.  We use a convolutional kernel with widths from 1 to 7, which covers up to 7 character n-grams (Figure~\ref{fig:charcnn}).  This allows the model to automatically learn lexical representations based on prefix, suffix, and stem information about a word.  No additional changes to learning are required for extension.

Adding the convolution does not dramatically slow down our model because the emission distributions can be computed for the whole batch in one operation. We simply pass the whole vocabulary through the convolution in a single operation.

\section{Infinite Context with LSTMs}
\label{sec:lstm}

One of the most powerful strengths of neural networks is their ability to create compact representation of data.  We will explore this here in the creation of transition matrices.  
In particular, we chose to augment the transition matrix with all preceding words in the sentence: $p(z_t \given z_{t-1}, w_0, \dots , w_{t-1})$.  Incorporating this amount of context in a traditional HMM is intractable and impossible to estimate, as the number of parameters grows exponentially.  

\begin{figure}
\centering
\includegraphics[width=0.66\linewidth]{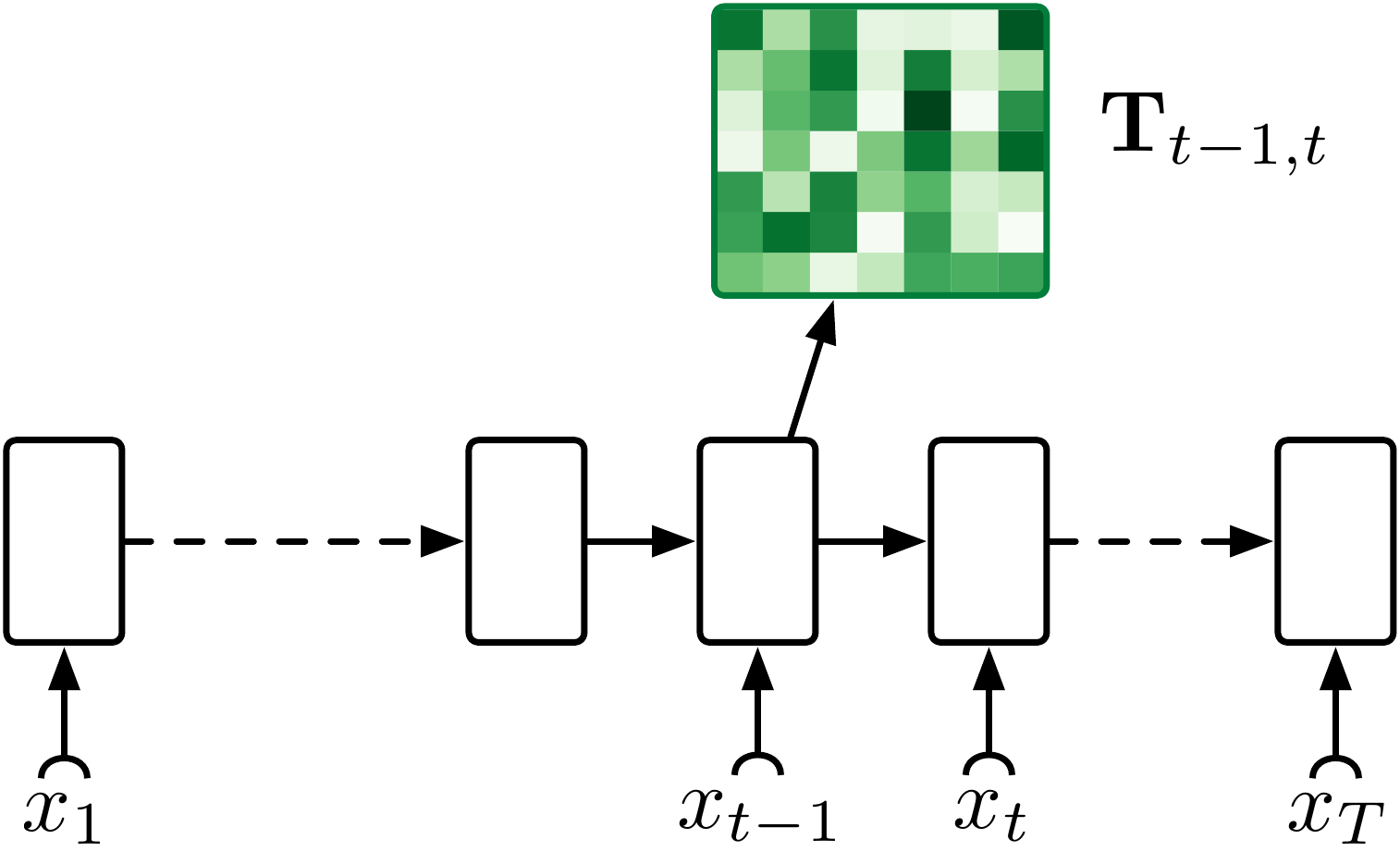}
\caption{A graphical representation of our LSTM transition network. Transition matrix $\mathbf{T}_{t-1,t}$ from time step $t-1$ to $t$ is computed based on the hidden state of the LSTM at time $t-1$.}
\label{fig:lstmtrans}
\end{figure}

For this reason, we use a stacked LSTM to form a low dimensional representation of the sentence ($C_{0\dots t-1}$) which can be easily fed to our network when producing a transition matrix: $p(z_t \given z_{t-1}, C_{0\dots t-1})$ in Figure \ref{fig:lstmtrans}.  By having the LSTM only consume up to the previous word, we do not break any sequential generative model assumptions.\footnote{This interpretation does not complicate the computation of forward-backward messages when running Baum-Welch, though it does, by design, break Markovian assumption about knowledge of the past.}  In terms of model architecture, the query embedding $\mathbf{q}$ will be replaced by a hidden state $\mathbf{h}_{t-1}$ of the LSTM at time step $t-1$.

\section{Evaluation}
Once a model is trained, the one best latent sequence is extracted for every sentence and evaluated on three metrics.
\paragraph{Many-to-One (M-1)}
Many-to-one computes the most common true part-of-speech tag for each cluster.  It then computes tagging accuracy as if the cluster were replaced with that tag.  This metric is easily gamed by introducing a large number of clusters.
\paragraph{One-to-One (1-1)}
One-to-One performs the same computation as Many-to-One but only one cluster is allowed to be assigned to a given tag.  This prevents the gaming of M-1.
\paragraph{V-Measure (VM)}
V-Measure is an F-measure which trades off conditional entropy between the clusters and gold tags.  \newcite{Christodoulopoulos:2010wv} found VM is to be the most informative and consistent metric, in part because it is agnostic to the number of induced tags.

\section{Data and Parameters}
To evaluate our approaches, we follow the existing literature and train and test on the full WSJ corpus. There are three components of our models which can be tuned.  Something we have to be careful of when train and test are the same data.  To avoid cheating, no values were tuned in this work.

\paragraph{Architecture}
The first parameter is the number of hidden units.  We chose 512 because it was the largest power of two we could fit in memory.  When we extended our model to include the convolutional emission network, we only used 128 units, due to the intensive computation of Char-CNN over the whole vocabulary per minibatch. 

The second design choice was the number of LSTM layers.  We used a three layer LSTM as it worked well for \cite{Tran:2016}, and we applied dropout \cite{srivastava:2014} over the vertical connections of the LSTMs \cite{pham:2013} with a rate of 0.5.  

Finally, the maximum number of inner loop updates applied per batch is set to six. We train all the models for five epochs and perform gradient clipping whenever the gradient norm is greater than five. To determine when to stop applying the gradient during training we simply check when the log probability has converged ($\frac{\text{new}-\text{old}}{\text{old}} < 10^{-4}$) or if the maximum number of inner loops has been reached.  All optimization was done using Adam \cite{Kingma:2015adam} with default hyper-parameters.

\paragraph{Initialization}
In addition to architectural choices we have to initialize all of our parameters.
Word embeddings (and character embeddings in the CNN) are drawn from a Gaussian $\mathcal{N}(0, 1)$. The weights of all linear layers in the model are drawn from a uniform distribution with mean zero and a standard deviation of $\sqrt{\nicefrac{1}{n_{\text{in}}}}$, where $n_{\text{in}}$ is the input dimension of the linear layer.\footnote{This is the default parameter initialization in Torch.} Additionally, weights for the LSTMs are initialized using $\mathcal{N}(0, \nicefrac{1}{2n})$, where $n$ is the number of hidden units, and the bias of the forget gate is set to 1, as suggested by \newcite{jozefowicz:2015}.  We present some parameter and modeling ablation analysis in \S \ref{parameter-ablation}.

It is worth emphasizing that parameters are shared at the lower level of our network architectures (see Figure~\ref{fig:charcnn} and Figure~\ref{fig:lstmtrans}). Sharing parameters not only allows the networks to share statistical strength, but also reduces the computational cost of computing sufficient statistics during training due to the marginalization over latent variables.

In all of our experiments, we use minibatch size of 256 and sentences of 40 words or less due to memory constraints. Evaluation was performed on all sentence lengths.  Additionally, we map all the digits to $0$, but do not lower-case the data or perform any other preprocessing. 
All model code is available online for extension and replication at\\
{\small \href{https://github.com/ketranm/neuralHMM}{\nolinkurl{https://github.com/ketranm/neuralHMM}}}.

\section{Results}
\begin{table}
\centering
\begin{tabular}{@{}l@{\hspace{5pt}}lc@{\hspace{5pt}}c@{\hspace{5pt}}c@{}}
\toprule
& System & M-1 & 1-1 & VM \\
\midrule
\multirow{2}{*}{\rotatebox[origin=c]{90}{Base}}
& HMM   & 62.5 & 41.4 & 53.3 \\
& Brown & 68.2 & 49.9 & 63.0 \\
\midrule
\multirow{4}{*}{\rotatebox[origin=c]{90}{SOTA}}
& Clark (2003)    & 71.2 & & 65.6\\
& Christodoulopoulos (2011) & 72.8 & & 66.1\\
& Blunsom (2011)  & \bf 77.5 & &\bf 69.8 \\
& Yatbaz (2012)   & 80.2 & & 72.1  \\
\midrule
\multirow{4}{*}{\rotatebox[origin=c]{90}{Our Work}}
& NHMM           & 59.8 & 45.7 & 54.2 \\
& + Conv         & 74.1 & 48.3 & 66.1 \\
& + LSTM         & 65.1 & 52.4 & 60.4 \\
& + Conv \& LSTM &\bf 79.1 &\bf 60.7 &\bf 71.7 \\
\bottomrule
\end{tabular}
\caption{English Penn Treebank results with 45 induced clusters. We see significant gains from both morphology (+Conv) and extended context (+LSTM).  The combination of these approaches results in a very simple system which is competitive with the best generative model in the literature.}
\label{tab:English}
\end{table}

Our results are presented in Table \ref{tab:English} along with two baseline systems, and the four top performing and state-of-the-art approaches.   As noted earlier, we are happy to see that our NHMM performs almost identically with the standard HMM.  Second, we find that our approach, while simple and fast, is competitive with Blunsom (2011).  Their Hierarchical Pitman-Yor Process for trigram HMMs with character modeling is a very sophisticated Bayesian approach and the most appropriate comparison to our work.

We see that both extended context (+LSTM) and the addition of morphological information (+Conv) provide substantial boosts to performance. Interestingly, the gains are not completely complementary, as we note that the six and twelve point gains of these additions only combine to a total of sixteen points in VM improvement.  This might imply that at least some of the syntactic context being captured by the LSTM is mirrored in the morphology of the language.  This hypothesis is something future work should investigate with morphologically rich languages.

Finally, the newer work of \newcite{yatbaz-sert-yuret:2012:EMNLP-CoNLL} outperforms our approach.  It is possible our performance could be improved by following their lead and including knowledge of the future.

\section{Parameter Ablation}
\label{parameter-ablation}

\begin{table}
\centering
\begin{tabular}{@{}l  c c c@{}}
\toprule
Configuration & M-1 & 1-1 & VM \\
\midrule
Uniform initialization    & 65.5 & 50.1 & 61.7\\
1 LSTM layer, no dropout  & 69.3 & 52.7 & 63.6 \\
1 LSTM layer, dropout     & 71.0 & 55.7 & 66.2\\
3 LSTM layers, no dropout & 72.7 & 52.2 & 65.1 \\
\midrule
Best Model &\bf 79.1 &\bf 60.7 &\bf 71.7 \\
\bottomrule
\end{tabular}
\caption{Exploring different configurations of NHMM}
\label{tab:conf}
\end{table}

Our model design decisions and weight initializations were chosen based on best practices set forth in the supervised training literature. We are lucky that these also behaved well in the unsupervised setting. Within unsupervised structure prediction, to our best knowledge, there has been no empirical study on neural network architecture design and weight initialization. We therefore provide an initial overview on the topic for several of our decisions.

\paragraph{Weight Initialization} If we run our best model (NHMM+Conv+LSTM) with all the weights initialized from a uniform distribution $\mathcal{U}(-10^{-4}, 10^{-4})$\footnote{We choose small standard derivation here for numerical stability when computing forward-backward messages.} we find a dramatic drop in V-Measure performance (61.7 vs 71.7 in Table~\ref{tab:conf}).  This is consistent with the common wisdom that unlike supervised learning \cite{luong:EMNLP}, weight initialization is important to achieve good performance on unsupervised tasks.  It is possible that performance could be further enhance via the popular technique of ensembling, would allow for combining models which converged to different local optima.

\paragraph{LSTM Layers And Dropout} We find that dropout is important in training an unsupervised NHMM.  Removing dropout causes performance to drop six points. To avoid tuning the dropout rate, future work might investigate the effect of variational dropout \cite{kingma:2015nisp} in unsupervised learning. We also observed that the number of LSTM layers has an impact on V-Measure.  Had we simply used a single layer we would have lost nearly five points. It is possible that more layers, perhaps coupled with more data, would yield even greater gains.

\section{Future Work}
In addition to parameter tuning and multilingual evaluation, the biggest open questions for our approach are the effects of additional data and augmenting the loss function.  Neural networks are notoriously data hungry, indicating that while we achieve competitive results, it is possible our model will scale well when run with large corpora.  This would likely require the use of techniques like NCE \cite{gutmann:2010} which have been shown to be highly effective in related tasks like neural language modeling \cite{Mnih:2012,vaswani:2013}.  Secondly, despite focusing on ways to augment an HMM, Brown clustering and systems inspired by it perform very well.  They aim to maximize mutual information rather than likelihood.  It is possible that augmenting or constraining our loss will yield additional performance gains.

Outside of simply maximizing performance on tag induction, a more subtle, but powerful contribution of this work may be its demonstration of the easy and effective nature of using neural networks with Bayesian models traditionally trained by EM.  We hope this approach scales well to many other domains and tasks.

\section*{Acknowledgments}
This work was supported by Contracts W911NF-15-1-0543 and HR0011-15-C-0115 with the US Defense Advanced Research Projects Agency (DARPA) and the Army Research Office (ARO). Additional thanks to Christos Christodoulopoulos.

\bibliography{emnlp2016}
\bibliographystyle{emnlp2016}

\end{document}